\documentclass[preprint,11pt,5p,twocolumn,authoryear]{elsarticle}
\usepackage[utf8]{inputenc}
\usepackage{amssymb}
\usepackage{bm}
\usepackage{mathtools}
\usepackage{xfrac}
\usepackage{graphicx}
\usepackage{caption}
\usepackage{subcaption}
\usepackage{etoolbox}

\usepackage[hidelinks]{hyperref}
\usepackage[nameinlink]{cleveref}

\usepackage[natbibapa]{apacite}

\newcommand{\matr}[1]{\mathbf{#1}}

\makeatletter
\providecommand{\doi}[1]{%
  \begingroup
    \let\bibinfo\@secondoftwo
    \urlstyle{rm}%
    \href{http://dx.doi.org/#1}{%
      doi:\discretionary{}{}{}%
      \nolinkurl{#1}%
    }%
  \endgroup
}

\let\oldhref\href
\renewcommand{\href}[2]{\oldhref{#1}{\hbox{#2}}}

\patchcmd\@combinedblfloats{\box\@outputbox}{\unvbox\@outputbox}{}{%
   \errmessage{\noexpand\@combinedblfloats could not be patched}%
}%
\makeatother

\begin{document}

\bibliographystyle{apacite}

\hyphenation{Conv-LSTM}
\hyphenation{Conv-LSTMs}

\begin{frontmatter}



\title{Multi-output Bus Travel Time Prediction\\
with Convolutional LSTM Neural Network}

\journal{Expert Systems with Applications}

\address[movia]{Public Transport Movia, Gammel Køge Landevej 3, 2500 Valby, Denmark}
\address[dtu]{Department of Management Engineering, Technical University of Denmark, 2800 Kongens Lyngby, Denmark}

\author[movia,dtu]{Niklas Christoffer Petersen\corref{cor1}}
\ead{niklch@dtu.dk}
\author[dtu]{Filipe Rodrigues}
\ead{rodr@dtu.dk}
\author[dtu]{Francisco Camara Pereira}
\ead{camara@dtu.dk}
\cortext[cor1]{Corresponding author}

\begin{abstract}
Accurate and reliable travel time predictions in public transport networks are essential for delivering an attractive service that is able to compete with other modes of transport in urban areas. The traditional application of this information, where arrival and departure predictions are displayed on digital boards, is highly visible in the city landscape of most modern metropolises. More recently, the same information has become critical as input for smart-phone trip planners in order to alert passengers about unreachable connections, alternative route choices and prolonged travel times. More sophisticated \textit{Intelligent Transport Systems} (ITS) include the predictions of connection assurance, i.e.\ an expert system that will decide to hold services to enable passenger exchange, in case one of the services is delayed up to a certain level. In order to operate such systems, and to ensure the confidence of passengers in the systems, the information provided must be accurate and reliable. Traditional methods have trouble with this as congestion, and thus travel time variability, increases in cities, consequently making travel time predictions in urban areas a non-trivial task. 
This paper presents a system for bus travel time prediction that leverages the non-static spatio-temporal correlations present in urban bus networks, allowing the discovery of complex patterns not captured by traditional methods. The underlying model is a multi-output, multi-time-step, deep neural network that uses a combination of convolutional and long short-term memory (LSTM) layers.

The method is empirically evaluated and compared to other popular approaches for link travel time prediction and currently available services, including the currently deployed model at Movia, the regional public transport authority in Greater Copenhagen. We find that the proposed model significantly outperforms all the other methods we compare with, and is able to detect small irregular peaks in bus travel times very quickly.
\end{abstract}


\end{frontmatter}
\section{Introduction}
One of the most visible applications of \emph{Intelligent Transport Systems} (ITS), within the field of public transportation, is the display of real-time traffic information. This has happened, traditionally, in the form of arrival and departure times on digital departure boards at stops and stations, and more recently in smart-phone apps and in-vehicle infotainment screens. It is widely deployed in most major cities, and is now considered a standard method to deliver an attractive and competitive public transport service. To an  increasing extent, real-time information channels constitute the only source of passenger information.

Public transport authorities have long found that GPS trajectory data from already deployed \emph{Automatic Vehicle Location} systems (AVL) can be used in the production of arrival and departures times \citep{Tcrp48}.

Our motivation is to improve the accuracy yielded by current prediction methods by exploiting spatio-temporal correlations present in public transport networks. Our focus is especially on urban bus networks that often share considerable parts of the infrastructure with other modes of transport, and therefore are prone to ripple effects. The proposed system, and the information it produces, can be integrated into ITSs in various ways with different applications. In its most basic form, the system can simply substitute current methods as a data source in passenger information systems, presenting real-time arrival and departure times. Passengers presented with reliable travel times can make use of this information in their decision-making \citep{Cats2011}, e.g.\ choose alternative routes or modes of transport to avoid prolonged travel time on their current route. The availability of the information produced by the system to awaiting passengers can also simply function as a comforting assurance, as studies have shown that reliable real-time information at bus stops has a statistically significant dampening effect on the perceived waiting time \citep{Fan2016}.

A more \textit{intelligent} use of the information can be in the context of automated trip planners.  These already accept this kind of real-time information, e.g.\ using the General Transit Feed Specification (GTFS). This allows for alerting passengers or proposing alternative routes earlier on the passenger's trip when a connecting service might be unreachable due to prolonged travel times.

Operating more sophisticated ITS applications successfully requires, to an even larger extent, accurate and reliable travel time predictions, since the cost of making erroneous decisions based on the predictions increases. \citet{Lo2012} present a decision support system for bus holding that requires accurate estimated arrival times to function optimally. Other advanced ITS examples include \emph{connection assurance} between two low frequency public transport services, where an expert system advises the driver or the traffic management system that one of the services should wait for the other, based on the arrival time predictions for both services. If the travel time predictions are too optimistic, the expert system ends up advising to hold the connecting service for longer than anticipated, introducing a prolonged delay for the other service and the passengers already present on that service. The use of our proposed system in this context can be achieved with a simple rule-based decision engine on top of the travel time model presented in detail in this paper.

\subsection{Bus travel time prediction}
\begin{figure}[!ht]
  \center  
  \includegraphics[width=\columnwidth]{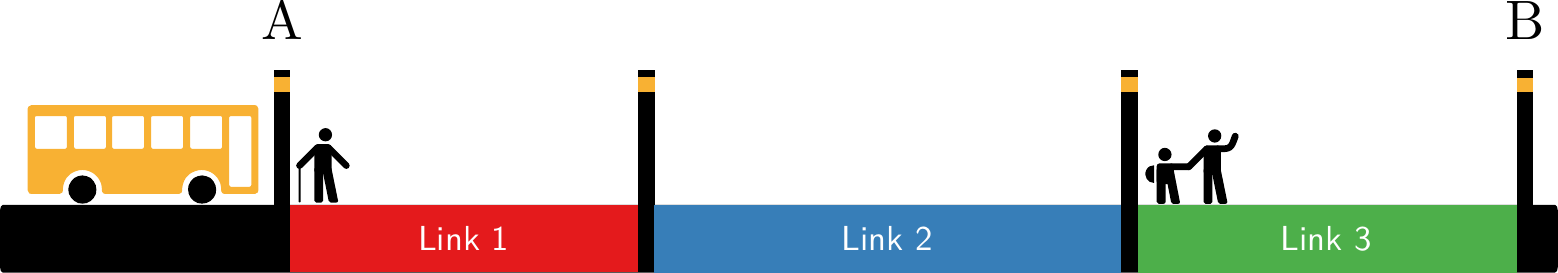}
  \caption{Arrival- and departure using link travel time.}
  \label{fig:bus-arrival-links}
\end{figure}

Arrival/departure time prediction is commonly approached as a specialization of travel time prediction as illustrated in \Cref{fig:bus-arrival-links}. The predicted travel time for each link is simply accumulated downstream the route to yield the arrival/departure time predictions at each stop point of the rest of the current journey. Thus, in the example, to predict when the bus at stop A will arrive at stop B, we would just sum up our predicted link travel times for Links 1 to 3. Besides the link travel time, estimations of dwell time (i.e.\ when a bus is holding at a stop point) should also be accumulated downstream.

Producing precise bus travel time predictions in areas with little external influence, e.g.\ rural areas, can to a large extent be solved with historical averaging or simple regression methods \citep{Williams2003,Altinkaya2013}. The problem becomes much more complex in urban areas where congestion, special events, roadworks, weather, etc.\ highly influence the traffic flow and passenger demand. As on-board GPS and AVL systems have become more affordable and common, data has both grown in coverage, i.e.\ number of vehicles with AVL installed, and frequency, i.e.\ number of GPS positions collected for each vehicle per time-unit.

Using geofencing techniques the raw GPS trajectory data can be converted into arrivals and departures at stop points, and subsequently, travel times on the links between the stop points. The objective is an intelligent expert system that utilizes this data in order to produce precise short-term predictions (e.g.\ up to\ $0-1.5$ hours in the future) for link travel time, specifically for bus traffic in urban areas.

Our contribution is an intelligent model for bus travel time prediction that takes advantage of the non-static spatio-temporal correlations present in urban bus traffic. We leverage on recent state-of-the-art techniques from \textit{machine learning} by combining  \emph{convolutional} and \emph{long short-term memory} (LSTM) \citep{Lstm1,Lstm2} neural networks, thus allowing the discovery of patterns across both time and space. Our proposed model is also multidimensional in its output with respect to both spatial and temporal aspects, i.e.\ we predict travel times for all links, for multiple time-steps ahead. 

The porposed method is empirically evaluated and compared to other popular approaches for link travel time prediction, including the model currently deployed in production by the public transport authority for the Greater Copenhagen Area, \emph{Movia}. Furthermore, the method is compared to Google Traffic (part of Google Maps), a popular online service for travel time prediction. 

This paper is structured in the following manner: In the next section, related work and relevant literature are reviewed. \Cref{sec:convlstm} introduces Convolutional LSTM neural networks in general, and in \Cref{sec:model} we present the proposed multi-output model in more detail, including e.g.\ network topology, and data preparation. \Cref{sec:experiments} introduces the Copenhagen dataset, which the model has been evaluated on, and our results are presented and discussed in~\Cref{sec:results}. Finally, we conclude on the work in~\Cref{sec:conclusion}.

\section{Related work}
Bus link travel time prediction has been explored in research as GPS and AVL data has become increasingly available. The problem overlaps with other research areas such as general traffic flow and speed estimation. But the problem has also unique constraints and opportunities that follow from servicing a fixed route with fixed stop points. The improvement in computational power in recent decades has gradually allowed more complex link travel time models with increased precision.

Early approaches for bus travel time prediction rely on historical average models \citep{Dailey1999,Sun2007}, and linear regression \citep{Patnaik2004}. Recent research presents this type of models only for comparison purposes, and in all cases, these are outperformed by the proposed alternatives \citep{Shalaby2004,Jeong2005}. The major disadvantage of historical average models is that they will only slowly converge to changes in the travel time, which of course is undesired with short, but highly impacting, external influences (e.g.\ a traffic incident or a large event). However, their simplicity, both with respect to computational cost and need of input data, has made them widely used in the industry. In rural areas, where traffic patterns are quite static, they can actually perform reasonably.

By their capabilities of maintaining state between predictions, Kalman filters (KF) have been the topic of several studies either as an independent model \citep{Chen2001,Shalaby2004}, or in combination with other models \citep{Yo2010,Bai2015}. In all cases, the applied filters are traditional linear KFs, and applied independently to each link. Because of the linearity, these models are computationally still quite cheap, but likewise, their disadvantage is that they are very limited in capturing and forecasting the complex non-linear dynamics of travel and dwell time in a metropolitan bus system. For example, the KF's state is only directly accessible for the leading time-step and thus is not capable of finding long-distance patterns spanning over several links and/or over several time-steps. In order to overcome this, KFs can be generalized to \textit{extended Kalman filters} (EKFs), allowing nonlinearities, but they still do not consider multiple links simultaneously. Making EKFs output travel times for all links simultaneously, with possible nonlinear interactions between them, would dramatically increase the computational cost.

The above analysis is substantiated by \citet{Lin2013} and \citet{Kumar2014} who find \textit{artificial neural networks} (ANN) to outperform independent Kalman filter models. However, the computational challenges of fully connected ANNs are also limiting the number of neurons of the network, and thus the complexity of the patterns it can learn to recognize. This has sparked the interest in studying \textit{composite} or \textit{hybrid} models. \citet{Bai2015} use a two-stage approach by combining an offline ANN model with an adaptable/online Kalman filter to yield a dynamic model. The advantage is the balance between computational complexity and the ability to adapt to smaller deviations quickly. The model is able to adapt to temporal variations in the current travel time on a journey, but it is still not able to recognize long distance patterns. The model proposed in this work uses \emph{long short-term memory} cells (LSTM), a apecial form of recurrent neural network (RNN) cells. \citet{Ma2015} use LSTM cells for highway speed prediction, and find it to significantly outperform KFs. Our proposal differs from existing research in bus link travel time prediction by combining the capability for maintaining state-space over multiple time-steps, while allowing the deep neural network to be efficiently trained.

Some recent research recognizes that several routes can benefit from each other's predictions if they share some partial route segment, e.g.\ \citep{Yu2011,Gal2014,Bai2015}. However, none of these approaches consider cross-temporal correlations between different route segments, and they only use a small window for correlation with upstream links (e.g.\ max.\ 3 links). Likewise, \citet{YanjieDuan2016} propose the use of an LSTM model for general highway travel time prediction, and to predict multiple time-steps ahead, but only for a single link at a time, i.e.\ cross link (spatial) correlations are lost. Another non-public transport study estimates travel times on road segments \citep{Tang2018}, and actually incorporates the spatial correlation, but the temporal aspect is very coarse and does not predict multiple time-steps ahead. In contrast, the combination of both \emph{LSTM} cells and \emph{convolutional} filters for bus travel time prediction, proposed in this paper, allows the learned patterns to generalize beyond a single link and time, i.e.\ multi-output and multi-time-step. Furthermore, this reduces the computational complexity by orders of magnitude compared to fully connected ANNs capable of capturing similar complex patterns. 

We can identify the following strengths of the proposed system compared to existing approaches:
\begin{enumerate}[i)]
    \item Unlike previous contributions in bus arrival prediction, it has the ability to learn spatio-temporal correlations as a coherent structure. The learned patterns can generalize over time and network links since the \emph{convolutional} filters are shared.
    \item We predict multiple time-steps ahead using a recurrent structure and an encoder-decoder architecture that allows the time-steps ahead to follow more complex patterns compared to existing approaches that just use a fully connected ANN layer as the final layer to split the prediction into multiple output time-steps.
    \item The input data needed for the method is easily obtained from the raw GPS traces that the AVL systems output, given the relatively fixed road network and location of bus stop points and stations.
\end{enumerate}

In contrast, the following possible weakness should also be considered:
\begin{enumerate}[i)]
    \item The computational complexity of the training is still a concern. Even though the computational complexity is reduced greatly with \emph{convolutional} filters compared to pure ANN models, it is still time-consuming to train the proposed model. That said, we have successfully trained models for complete routes using commodity-grade hardware within reasonable time. With our test setup we could do retraining on a daily basis without computational complications. The training can easily be distributed across multiple computational instances, so we argue that the scalability issue can be overcome.
\end{enumerate}

\section{Convolutional LSTM neural networks}
\label{sec:convlstm}
A long short-term memory (LSTM) neural network is a special type of Recurrent Neural Network~(RNN) which has been proven robust for capturing long-term dependencies \citep{Lstm1,Lstm2}. The important feature of an LSTM network is its capability to maintain a cell state, $\matr{c}_t$, from previous observations across sequences of input (e.g.\ time), but also to eliminate information considered irrelevant. To allow this mechanism, the maintenance of information is controlled by three gates: \emph{input gate}, \emph{forget gate}, and \emph{output gate}. Each gate yields a state variable at time $t$, respectively $\matr{i}_t$, $\matr{f}_t$, and $\matr{o}_t$, along with the cell output, $\matr{h}_t$, cf.~\cref{eq:lstm}, where $\circ$ denotes the element-wise product.

\begin{equation}
\begin{aligned}
  \matr{i}_t &= \sigma \left( \matr{W}^i \matr{x}_t + \matr{R}^i \matr{h}_{t-1} + \matr{U}^i \circ \matr{c}_{t-1} + \matr{b}^i \right) \\
  \matr{f}_t &= \sigma \left( \matr{W}^f \matr{x}_t + \matr{R}^f \matr{h}_{t-1} + \matr{U}^f \circ \matr{c}_{t-1} + \matr{b}^f \right) \\
  \matr{c}_t &= \matr{f}_t \circ \matr{c}_{t-1} + \matr{i}_t \circ \mathrm{tanh} \left( \matr{W}^c \matr{x}_t + \matr{R}^c \matr{h}_{t-1} + \matr{b}^c \right) \\
  \matr{o}_t &= \sigma \left( \matr{W}^\mathit{o} \matr{x}_t + \matr{R}^\mathit{o} \matr{h}_{t-1} + \matr{U}^\mathit{o} \circ \matr{c}_{t} + \matr{b}^o \right) \\
  \matr{h}_t &= \matr{o}_t \circ \mathrm{tanh} \left( \matr{c}_t \right)
\end{aligned}
\label{eq:lstm}
\end{equation}
\vspace{.5em}

\begin{figure}[!ht]
  \centering
  \includegraphics[scale=.45]{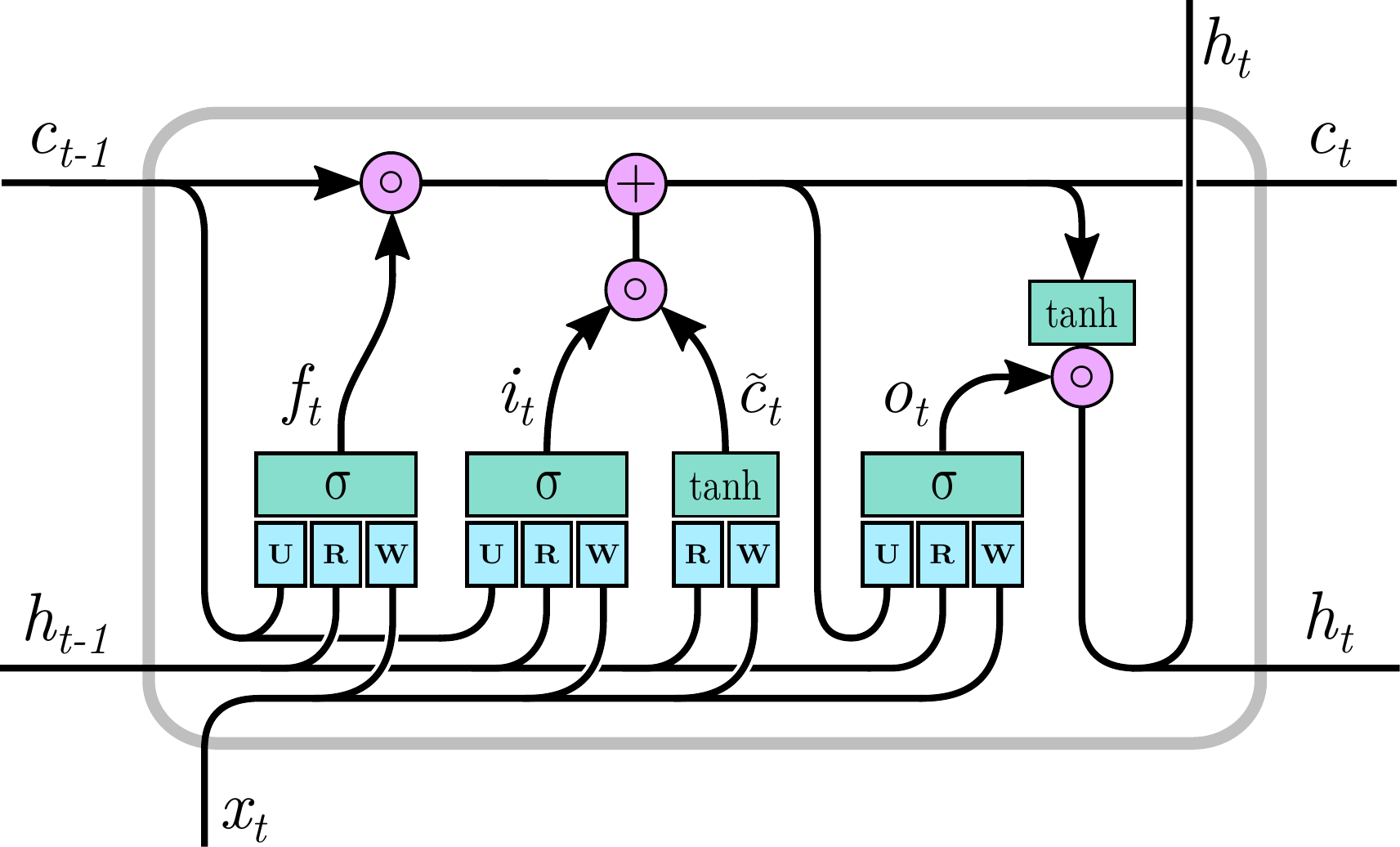}
  \caption{Structure of LSTM cell with peephole.}
  \label{fig:lstm-peepholes}
\end{figure}

\Cref{fig:lstm-peepholes} illustrates the inner structure of an LSTM cell with peephole as proposed by \cite{LstmPeephole}. 
It has especially grown popular for predicting time series using methods evolved from \cite{LstmTs}, where fixed-length windows of time-series are generated and feed into an LSTM network. Multiple LSTMs can be stacked such that more complex patterns of sequential information (e.g.\ temporal patterns) can be learned.

\begin{figure}[!ht]
  \centering
  \includegraphics[scale=.5]{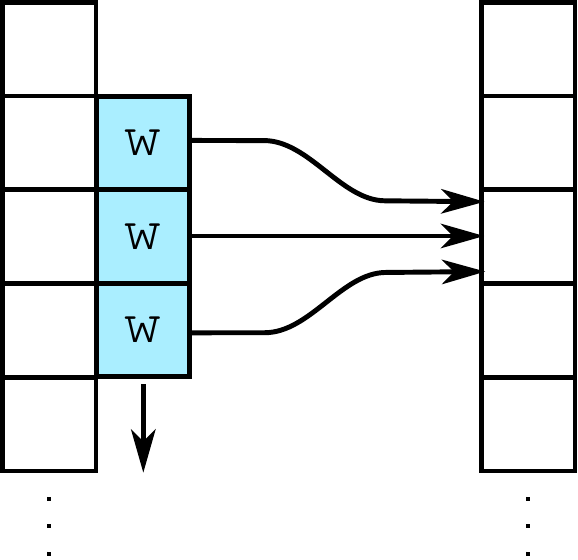}
  \caption{Application of convolutional filter onto 1D data.}
  \label{fig:conv}
\end{figure}

Convolutional Neural Networks~(CNNs), on the other hand, have been widely used for capturing spatial relationships, e.g.\ the importance of neighboring pixels in an image. As opposed to fully connected layers, where each unit $i$ in the layer has a dedicated scalar weight $w_{ij}$ for all input values $x_j$, convolutional units are only locally connected and reuse the same weights to produce several outputs. Instead of considering the entire input-vector, only a fixed-size window, or \emph{convolution}, around each input is considered. The weights are therefore referred to as the \emph{filters} or \emph{kernels} of the layer. \Cref{fig:conv} illustrates a single convolutional filter of size $3$ being applied to one-dimensional data.

Special care needs to be taken at the boundaries, i.e.\ where the convolutional filter will exceed the input. To avoid that the size of the output decreases, an approach is to pad the input, e.g.\ with zeros. This ensures that the output shape of each convolutional unit will always be identical to the input shape, which is often desirable. One of the key benefits of convolutional networks is that the number of weights that needs to be learned is considerably reduced compared to fully connected networks, and also that learned patterns can be transferred across space. I.e.,\ the convolutional filters become feature detectors that, in our case, can detect spatial patterns across links, e.g.\ congestion forming, etc.   

\citet{ConvLSTM} introduced the novel combination of convolutional and LSTM layers into a single structure, the \emph{Convolutional LSTM}, or simply \emph{ConvLSTM}. Specifically, the method applies convolutional filters in the \emph{input-to-state} and \emph{state-to-state} transitions of the LSTM cf.~\cref{eq:convlstm}, where $*$ denotes the convolution operator.
\begin{equation}
\begin{aligned}
  \matr{i}_t &= \sigma \left( \matr{W}^i * \matr{x}_t + \matr{R}^i * \matr{h}_{t-1} + \matr{U}^i \circ \matr{c}_{t-1} + \matr{b}^i \right) \\
  \matr{f}_t &= \sigma \left( \matr{W}^f * \matr{x}_t + \matr{R}^f * \matr{h}_{t-1} + \matr{U}^f \circ \matr{c}_{t-1} + \matr{b}^f \right) \\
  \matr{c}_t &= \matr{f}_t \circ \matr{c}_{t-1} + \matr{i}_t \circ \mathrm{tanh} \left( \matr{W}^c * \matr{x}_t + \matr{R}^c * \matr{h}_{t-1} + \matr{b}^c \right) \\
  \matr{o}_t &= \sigma \left( \matr{W}^\mathit{o} * \matr{x}_t + \matr{R}^\mathit{o} * \matr{h}_{t-1} + \matr{U}^\mathit{o} \circ \matr{c}_{t} + \matr{b}^o \right) \\
  \matr{h}_t &= \matr{o}_t \circ \mathrm{tanh} \left( \matr{c}_t \right)
\end{aligned}
\label{eq:convlstm}
\end{equation}
\vspace{.5em}

\begin{figure*}[!t]
  \centering
  \includegraphics[scale=.75]{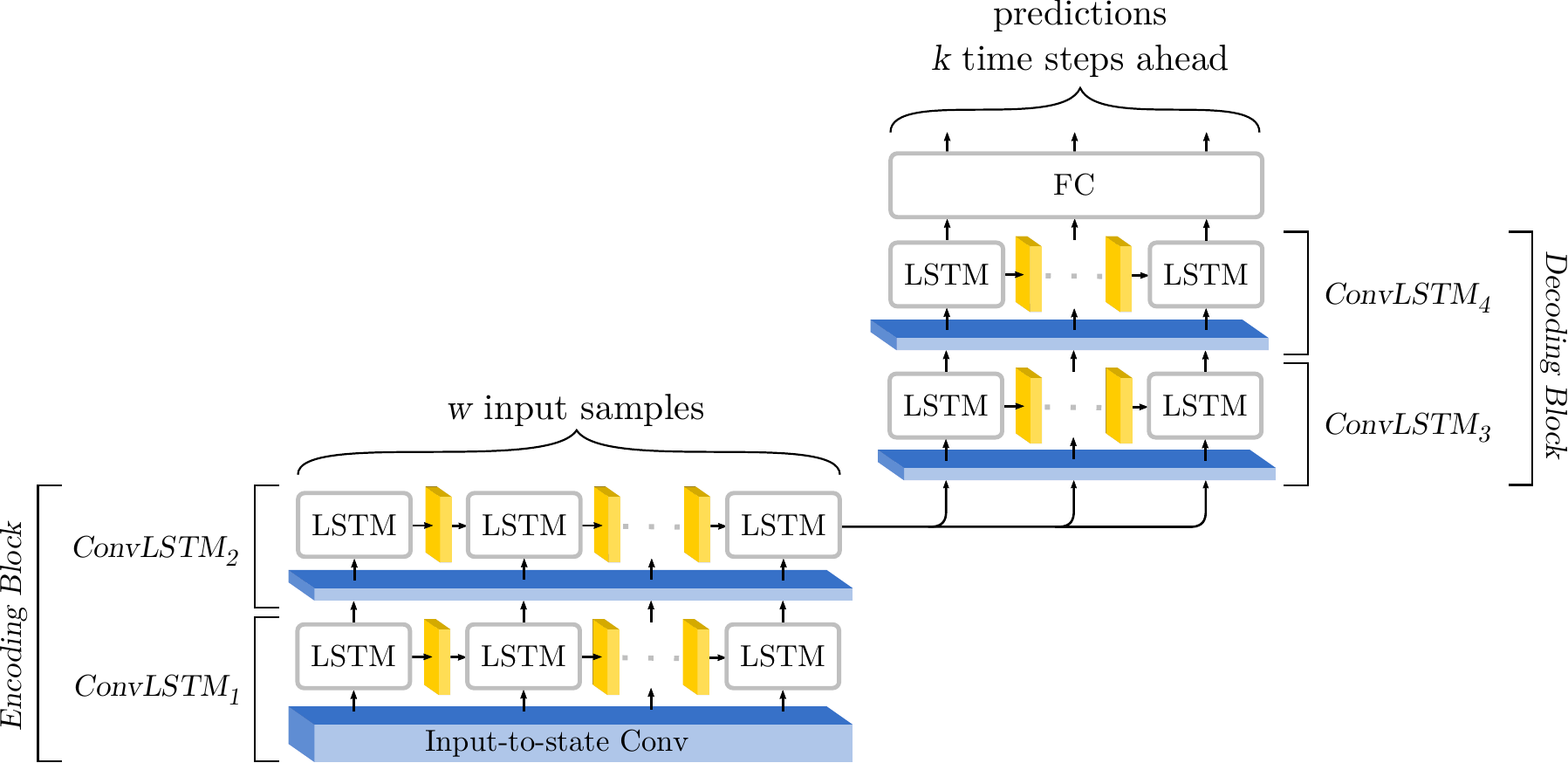}
  \caption{Convolutional LSTM network topology.}
  \label{fig:ConvLSTM}
\end{figure*}

As with traditional CNN layers, the output dimensionality of a \emph{ConvLSTM} layer is determined by the number of filters applied. However, \emph{ConvLSTMs} require a total of eight filters for each desired output, i.e.\ four \emph{input-to-state} filters ($\matr{W}^i$, $\matr{W}^f$, $\matr{W}^c$, and $\matr{W}^o$) and four \emph{state-to-state} filters ($\matr{R}^i$, $\matr{R}^f$, $\matr{R}^c$, and $\matr{R}^o$).
Still, it is important to emphasize that the application of convolutional filters to the LSTM model greatly reduces the number of parameters/weights that need to be learned, compared to a \emph{pure LSTM} approach. This allows for even deeper networks.
\vspace{-.3em}

\section{Multi-output model}
\label{sec:model}
In this section, we present the multi-output, multi-time-step model for bus travel time prediction that uses the \emph{ConvLSTM} layer introduced in the previous section.

\vspace{-.2em}
\subsection{Network topology}

\Cref{fig:ConvLSTM} shows the overall network topology, where blue boxes illustrate \emph{input-to-state} convolutions and yellow boxes \emph{state-to-state} convolutions. The network uses a sequence encoder/decoder technique, which is an extension of the encoder/decoder presented by \citet{ConvLSTM}. The encoder block consists of two \emph{ConvLSTM} layers, where the resultant sequence (last $k$ values of the sequence) is fed into a decoder, or prediction block. The decoder block also consists of two \emph{ConvLSTM} layers, and a fully connected (FC) layer. The proposed architecture allows unequal $w$ and $k$, e.g.\ it predicts the next $3$ time-steps based on a window size of $20$ previous time-steps.

Therefore, convolutional filters are applied to each input, at each time-step, to the respective LSTM cell and also between LSTM cells in the state-transition. Since the time-steps are one-dimensional~(i.e.\ link travel times across links), the filters are also one-dimensional. In each of the two blocks, the \emph{ConvLSTMs} are arranged with filter sizes of respectively $10\times1$ and $5\times1$ for each of the layers in the block. This size is used both for the \emph{input-to-state} and \emph{state-to-state} convolutional filters. Lastly, each \emph{ConvLSTM} layer has 64~outputs, yielding a total need of 512 convolutional filters.

In order to avoid over-fitting during training \emph{Dropout} \citep{Dropout} is used between the \emph{ConvLSTM} layers, and \emph{Batch Normalization} \citep{BatchNorm} is also performed before each \emph{ConvLSTM} layer to ensure reasonable inputs for the activations and speed-up learning. The dropout probability is adjusted to 20\%, 10\%, and 10\%, respectively.

Each of the \emph{ConvLSTM} layers uses linear activation functions, and the output from the last layer in the decoder block is fed into a fully connected (FC) layer using the \emph{ReLU} activation function, which also ensures that only positive travel times are predicted.

\subsection{Data preparation}
We expect link travel times from AVL systems to be available in a tabular form, where each link travel time measurement has a timestamp, and a reference to the link as illustrated in~\Cref{tab:data}. This output is standard for most AVL systems used in public transport systems, thus allowing the proposed system to generalize to other regions.
\begin{table}[!ht]
  \centering
  \footnotesize
  \begin{tabular}{llr}
    Timestamp & Linkref. & Link travel time (s) \\ \hline \hline
    2017-10-10 00:20:02 & 29848:1254 & 63 \\ \hline
    2017-10-10 00:21:07 & 1254:1255  & 65 \\ \hline
    2017-10-10 00:21:51 & 1255:10115 & 44 \\ \hline
    \vdots & \vdots & \vdots 
  \end{tabular}
  \caption{Example of raw travel time measurements.}
  \label{tab:data}
\end{table}

\begin{figure}[!ht]
  \centering
  \includegraphics[scale=1.0]{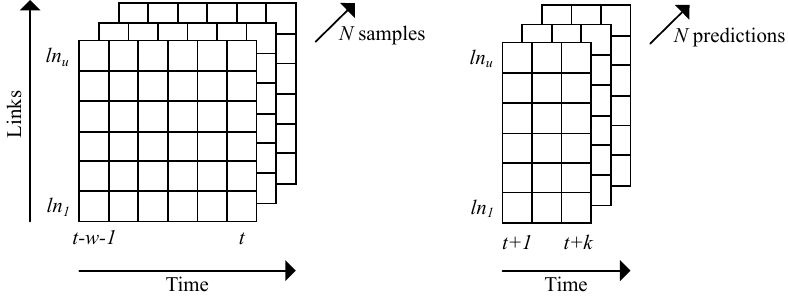}
  \caption{Shapes of the input and output data.}
  \label{fig:data_shape}
\end{figure}

For the \emph{ConvLSTM} model to be able to capture the desired spatio--temporal patterns, the input data must be arranged in a suitable manner, i.e.\ in $N$ samples, each with a window of the $w$ lagging time-steps $t-w+1, \ldots, t$, and each time-step with $u$ link travel times $1, \ldots, u$, as illustrated in~\Cref{fig:data_shape}.

As for the output, it consists of $N$ predictions for each of the $k$ time-steps ahead, $t+1, \ldots, t+k$. Thus the input is a 4D-tensor, $\matr{X}$ with dimensionality $N \times w \times u \times 1$, and the output, $\matr{Y}$, a 4D-tensor with dimensionality $N \times k \times u \times 1$. In both cases, the last one refers to the single link travel time for each time-step/link combination. It is emphasized that each prediction consists of travel time predictions for all links for the next $k$ time-steps, i.e.\ multi-output, multi-time-step-ahead prediction.

The $N$ samples are sampled at a fixed time resolution since we need a shared time reference across all links. \Cref{sec:experiments} elaborates on some of the considerations for choosing an adequate resolution.

\subsection{Detrending}
Urban bus travel times vary throughout the day, and the day of the week due to \emph{recurring congestion}. In order to reduce the need for the deep neural network to learn this recurring variation, the travel times for link $\mathit{ln} \in \{ 1,\ldots,u\}$, at time-step $t$, $x_{\mathit{ln},t}$, are normalized to focus on deviations from the normal and expected pattern. Travel times are centered with the mean for each link, at the time of day, and day of week, $\mathit{\bar{x}_{\mathit{ln},\mathit{dow},\mathit{tod}}}$, and scaled with the standard deviation for each link, $\sigma_\mathit{ln}$:

\begin{equation}
  x'_{\mathit{ln},t} = \frac{x_{\mathit{ln},t} - \bar{x}_{\mathit{ln},\mathit{dow},\mathit{tod}}}{\sigma_\mathit{ln}}
  \label{eq:normalization}
\end{equation}

A similar normalization is applied to the predicted travel times, $y_{\mathit{ln},t}$, but only using the historical mean and standard deviation, since the true mean and standard deviation are obviously unavailable in real-time prediction scenarios.

When calculating the mean and standard deviation, it can be beneficial to exclude extreme outliers, since both mean and standard deviations are highly sensitive to such measurements. A suggested method is to apply \emph{absolute deviation around the median} (MAD; see \cite{Olewuezi2011}) when calculating $\mathit{\bar{x}_{\mathit{ln},\mathit{dow},\mathit{tod}}}$ and $\sigma_\mathit{ln}$.

\subsection{Implementation and training}
The proposed network model was implemented in Python using the Keras Framework \citep{Keras}, and trained using the \emph{RMSprop} algorithm \citep{RMSprop}. The source code for the proposed method is publicly available at GitHub:  \cite{gh-bus-arrival-convlstm}. 

During training, the variables $\mathit{\bar{x}_{\mathit{ln},\mathit{dow},\mathit{tod}}}$ and $\sigma_\mathit{ln}$ should be calculated solely based on the training set, to emulate the real-world application.

\section{Experiments}
\label{sec:experiments}

For the purpose of evaluation, the proposed method is applied to a dataset from Copenhagen's public transport authority, \emph{Movia}. The dataset consists of 1,2M~travel time observations for the ``4A'' bus line in the period May to October 2017. The data points were collected using the real-time AVL system installed in every vehicle servicing the line. 

\begin{figure}[!ht]
  \centering
  \includegraphics[width=0.4\textwidth]{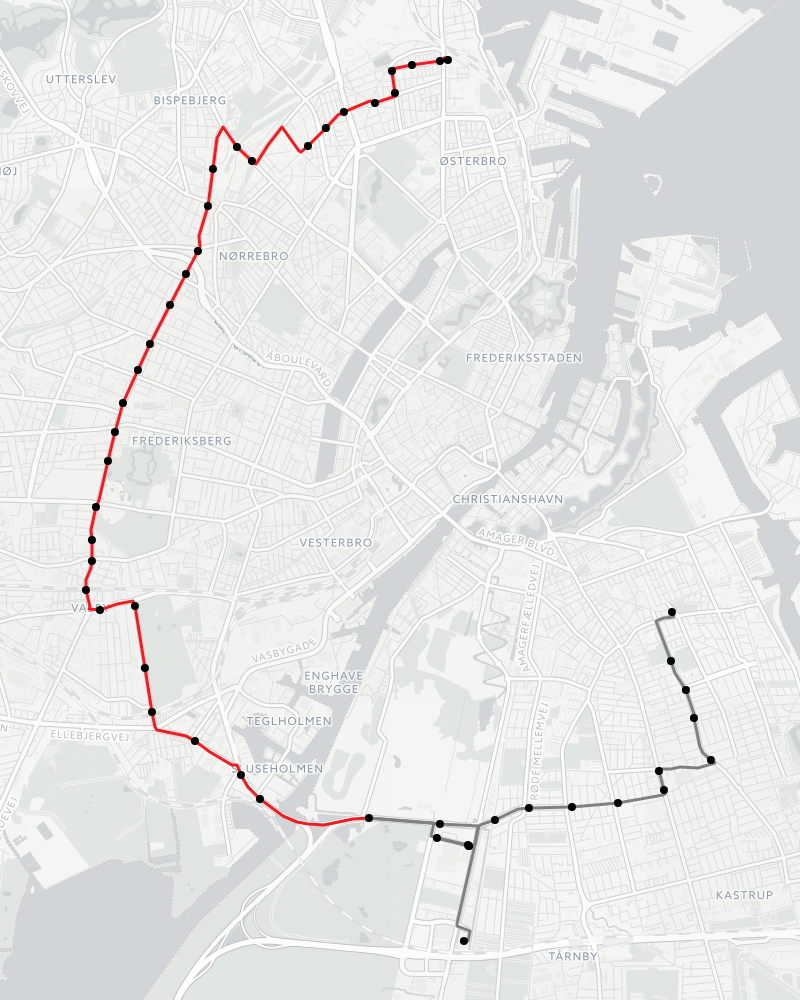}
  \caption{Geography of the 4A bus line in Copenhagen.}
  \label{fig:4a_map}
\end{figure}

The geography of the route is shown in \Cref{fig:4a_map}. As the line circles Central Copenhagen, it is highly sensitive to congestion to/from the city since it intersects with several large corridors along its route. Southeast of the city center, the line splits into different destination patterns (gray), therefore only the first 32 links are considered for the purposes of this experiment (red).

\begin{figure*}[!t]
    \centering
    \begin{subfigure}[t]{0.31\textwidth}
        \centering
        \includegraphics[width=\textwidth]{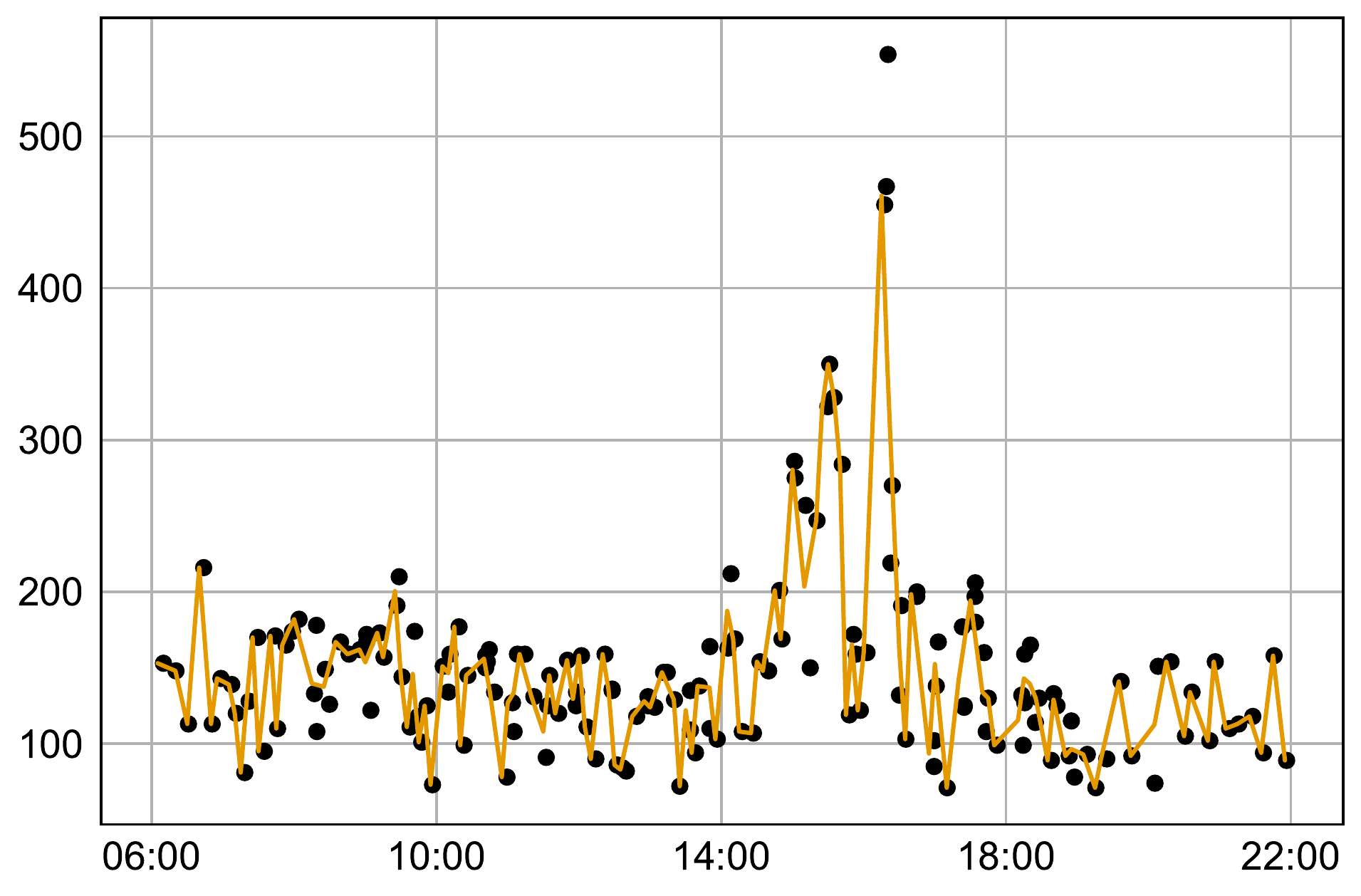}
        \caption{5 min}
    \end{subfigure}%
    ~ 
    \begin{subfigure}[t]{0.31\textwidth}
        \centering
        \includegraphics[width=\textwidth]{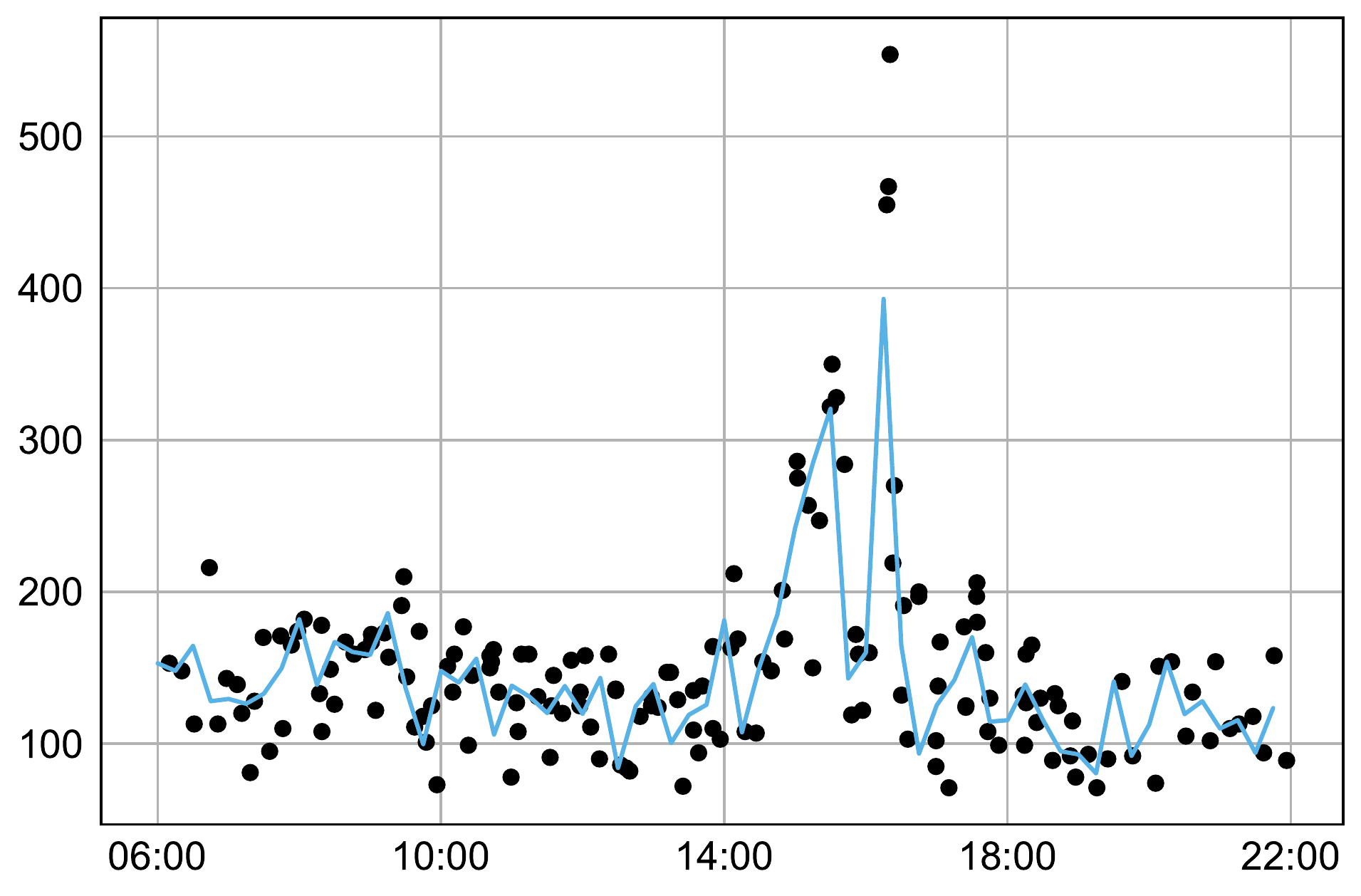}
        \caption{15 min}
    \end{subfigure}
    ~
    \begin{subfigure}[t]{0.31\textwidth}
        \centering
        \includegraphics[width=\textwidth]{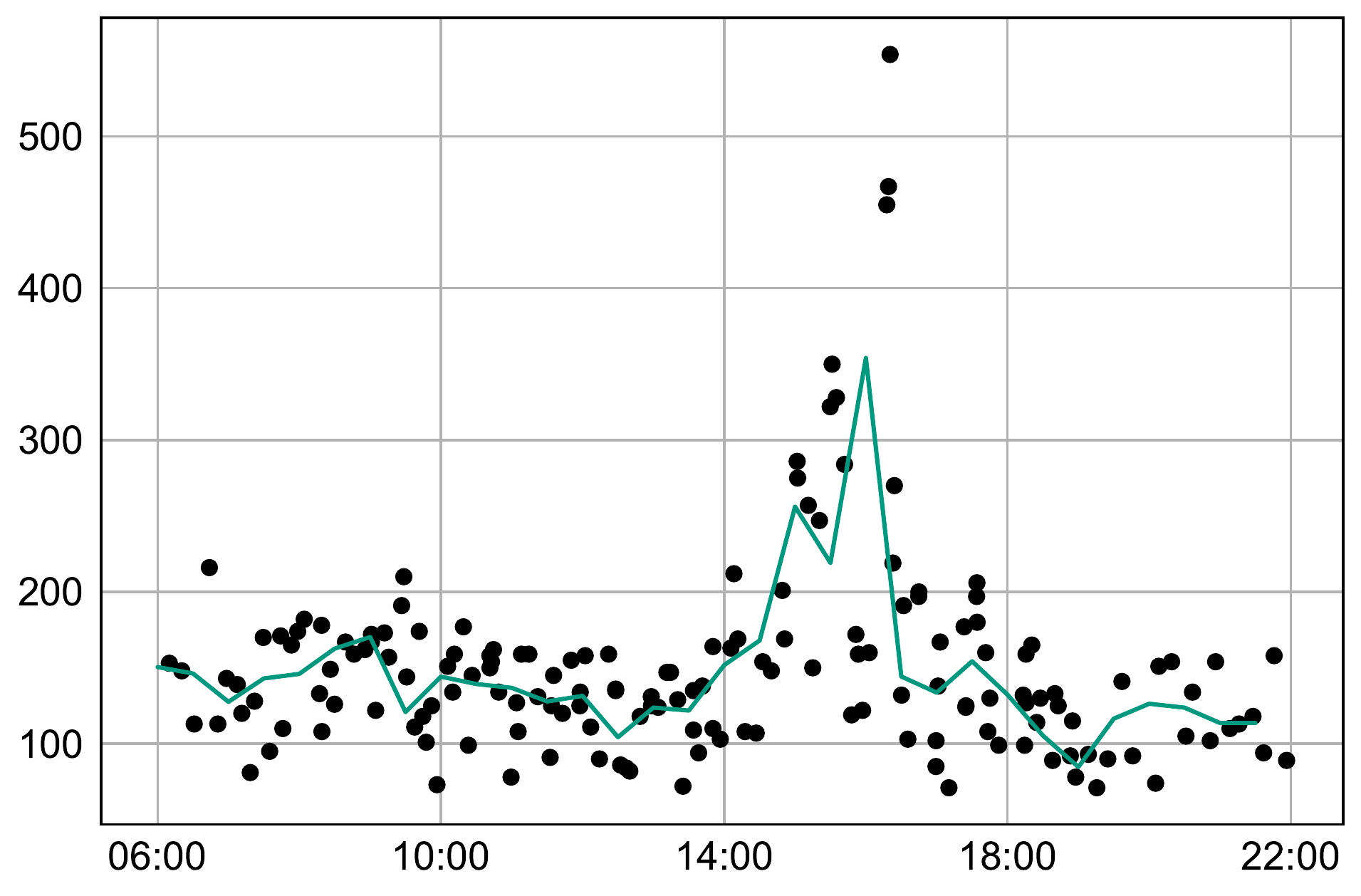}
        \caption{30 min}
    \end{subfigure}

    \caption{Examples of travel time for a single link over a single day, at various time resolutions.}
    \label{fig:resolutions}
\end{figure*}

\subsection{Time resolution}
In order to allow predictions for fixed time-steps ahead, the data is aggregated at a fixed time resolution. The choice of time resolution is a hyper-parameter for the proposed system, and should be tuned for the specific dataset. \Cref{fig:resolutions} shows examples of travel time for a single link over a single day at various time resolutions. The black dots are actual measurements, and the lines the aggregated mean link travel time at the given resolution. Several considerations should be made when choosing the time resolution:
\begin{itemize}
    \item The \emph{expected} frequency of the line, since a choice far from this will lead to either 1) sparse measurements, and low probability of actually using a prediction, because no service runs in the predicted time step; or 2) an overly smooth time-series, with too much detail about variability being lost. Thus it is a balance between capturing the details and still having a reasonable number of measurements of each time-step to avoid overfitting.  
    \item The computational cost of training the system, since smaller time-steps will require further iterations over the training data and larger values of $w$ and $k$ to include the same lagging time window, and time horizon for predictions. 
\end{itemize}

\begin{figure}[!ht]
    \centering
    \includegraphics[width=0.4\textwidth]{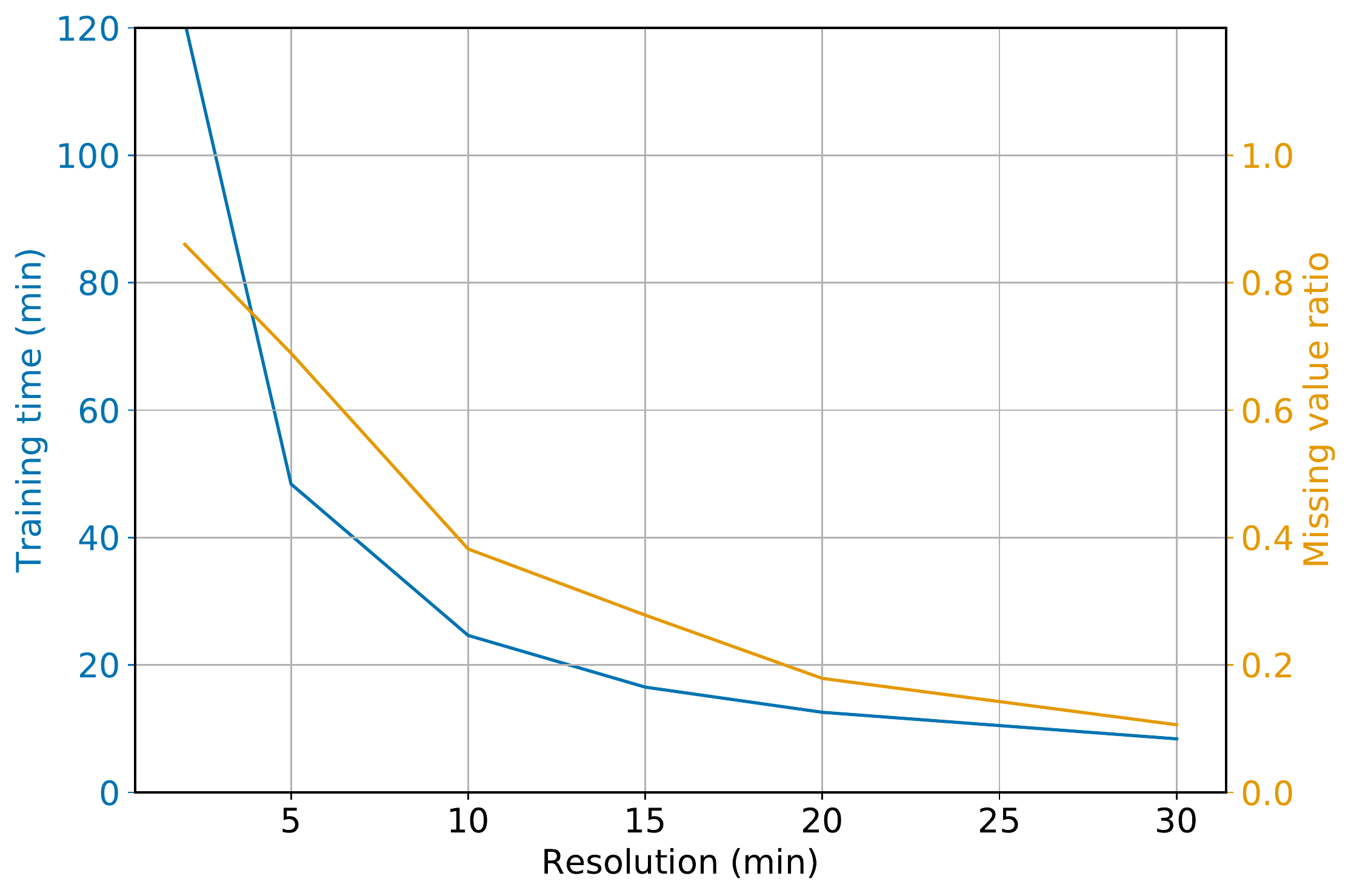}
    \caption{Choice of resolution influence on training time and missing values.}
    \label{fig:runtime_stats}
\end{figure}

\Cref{fig:runtime_stats} shows how the choice of resolution influences the training time of our proposed deep neural network architecture on commodity hardware (blue). It also shows how the portion of time-steps with missing values (yellow) also increases as more fine-grained resolutions are considered. For instance, using a 2-minute resolution will cause 89\% of all time steps to not include any measurements.

For this experiment, the AVL data was aggregated into 15-minute intervals and normalized as described in~\Cref{sec:model}. This resolution was chosen based on the above-mentioned considerations. The ``4A'' bus line had a measured mean \emph{headway} (the time between two vehicles during daytime) of $7.5$ minutes between 06:00 and 22:00, and thus there is a reasonably high probability that 15-minute time-steps will include 1-2 measurements. Indeed, the average number of measurements in each time step was $1.7$ for the training set.

Given the time resolution, we set the fixed window size, $w = 32$, equivalent to 8 hours, to allow patterns in the morning peak to affect patterns in the afternoon peak. We set $k = 3$ to allow predictions of up to 45 minutes into the future. 

\subsection{Evaluation}
The evaluation of the proposed model and all the considered baselines is based on the following statistics: \emph{mean absolute error} (MAE), \emph{root mean square error} (RMSE), and \emph{mean absolute percentage error} (MAPE), as formalized in \cref{eq:mae,eq:rmse,eq:mape}, where $\matr{Y}_i$ is the true link travel times for sample $i$ and $\matr{\widehat{Y}}_i$ is the predicted travel times. Since the multi-output, multi-time-step model predicts link travel times for all $u$ links for the next $k$ time-steps, $\matr{Y}_i$ and $\matr{\widehat{Y}}_i$ have the same dimensionality: $w \times u \times 1$.

\begin{equation}
    \textrm{MAE}(\matr{Y}, \matr{\widehat{Y}}) = \frac{\sum_{i = 1}^{N} \left| \matr{Y}_i - \matr{\widehat{Y}}_i \right| }{N}
    \label{eq:mae}
\end{equation}

\begin{equation}
    \textrm{RMSE}(\matr{Y}, \matr{\widehat{Y}}) = \sqrt{\frac{\sum_{i = 1}^{N} \left(\matr{Y}_i - \matr{\widehat{Y}}_i \right)^2}{N}}
    \label{eq:rmse}
\end{equation}

\begin{equation}
    \textrm{MAPE}(\matr{Y}, \matr{\widehat{Y}}) = \frac{1}{N} \sum_{i = 1}^{N} \left| \frac{\matr{Y}_i - \matr{\widehat{Y}}_i}{\matr{Y}_i} \right| 
    \label{eq:mape} 
\end{equation}
\vspace{.5em}

\begin{table*}[!t]
    \center
    \begin{tabular}{ll|rrr}
        Model & Time ahead & RMSE (min) & MAE (min) & MAPE (\%) \\
        \hline         
        \hline
        Historical average &                & 4.35 & 3.23 & 6.51 \% \\ 
        \hline 
        Current model      & t + 1 (15 min) & 4.92 & 3.90 & 8.05 \% \\
                           & t + 2 (30 min) & 4.91 & 3.46 & 6.82 \% \\
                           & t + 3 (45 min) & 5.47 & 4.15 & 8.68 \% \\
        \hline 
        Pure LSTM          & t + 1 (15 min) & 3.48 & 2.48 & 5.02 \% \\
                           & t + 2 (30 min) & 3.56 & 2.51 & 5.08 \% \\
                           & t + 3 (45 min) & 3.68 & 2.62 & 5.34 \% \\
        \hline 
        Google Traffic     & t + 1 (15 min) & 3.67 & 2.96 & 6.32 \% \\ 
        \hline 
        ConvLSTM           & t + 1 (15 min) & 2.66 & 1.99 & 4.19 \% \\
                           & t + 2 (30 min) & 2.89 & 2.11 & 4.44 \% \\
                           & t + 3 (45 min) & 3.11 & 2.27 & 4.75 \% \\
        \hline 
    \end{tabular}
    \caption{Results of the proposed and the baseline models}
    \label{tab:results}
\end{table*}

To allow a clear comparison, we reduce $\matr{Y}_i$ and $\matr{\widehat{Y}}_i$ by summing over all links:
\begin{align}
    \matr{Y}_i' = \sum_{\mathit{ln} = 1}^{u} \matr{Y}_{i,\mathit{ln}} & & \matr{\widehat{Y}}_i' = \sum_{\mathit{ln} = 1}^{u} \matr{\widehat{Y}}_{i,\mathit{ln}}
    \label{eq:total} 
\end{align}
This is equivalent to predicting the total travel time of all 32 links, and follows the initial approach for arrival/departure time prediction by accumulating link travel times.

The output of each of the evaluation functions is thus simply a vector of size $k$, i.e.\ the evaluation of the different time-steps for all links accumulated.

The model is trained on the prepared data using a sliding window approach in order to simulate real-world conditions, in which real-time travel time measurements arrive as a continuous data stream. We use 23 weeks of data for training, and one week of data for testing. The window is advanced for 1 week at a time for a total of 4 test weeks. The trained models are available, alongside the source code, at \citep{gh-bus-arrival-convlstm} and include a test dataset (4 weeks). For replicating our results, the full dataset used in this experiment is available from Movia upon request.

\section{Results and discussion}
\label{sec:results}
The performance of our proposed model for link travel time prediction, based on \emph{ConvLSTM}, is compared against several baseline models and services:
\begin{enumerate}
   \item a naïve historical average model, i.e.\ equivalent to just predicting the normalized value, $\bar{x}_{\mathit{ln},\mathit{dow},\mathit{tod}}$;
   \item the traffic prediction model currently deployed by Movia;
   \item a pure LSTM model for link travel time prediction, i.e.\ without applying convolutional filters in state transitions;
   \item travel time predictions from Google Traffic (part of Google Maps).
\end{enumerate}

\Cref{tab:results} shows the overall performance of the proposed model and the  baseline methods. Predictions are limited to daytime, i.e.\ between 06:00 and 22:00, and are accumulated downstream on a journey level to simulate the use for real-time bus arrival/departure time prediction, cf.~\cref{eq:total}. 

Before going into a direct comparison, it is important to understand some aspects of the baseline models, and how measurements were collected.

\subsection{Historical average}
The performance of the historical average is independent with respect to the number of time-steps ahead in time it predicts, as it just represents a weekly cycle of mean link travel times.

\subsection{Current model}
Measurements from the currently deployed bus prediction model were collected at a 5-minute frequency using a non-publicly accessible endpoint at the transport authority. The model is based on a historical average model, but also has a rule-based mechanism on top that can override or adjust the historical link travel times. For instance, it will, to some extent, assume that a delayed vehicle will recover from its delay by traversing links a bit faster. Of course, such an assumption can be problematic in an urban area with many external traffic effects. 

\subsection{Pure LSTM}
The pure LSTM model for link travel time prediction is similar to the model proposed by \citet{YanjieDuan2016}. The model was trained on the exact same dataset as the \emph{ConvLSTM} model and has a similar architecture, but without the convolutional filters.

\subsection{Google Traffic}
Measurements from the Google Traffic model were collected using the Google Maps Distance Matrix API~\citep{GMaps_DistanceMatrixAPI}. Google uses crowd-sourced road congestion data collected from smart-phones with the \emph{Google Maps App} installed \citep{GMaps_Crowdsourcing}. While the exact model powering the service is not publicly described in detail, the documentation states that ``the returned \emph{duration in traffic} should be the best estimate of travel time given what is known about both historical traffic conditions and live traffic''. Furthermore, it states that ``live traffic becomes more important the closer the departure time is to now'' \citep{GMaps_DistanceMatrixAPI}.

Because there is a limit to the number of requests that one can freely make to the API over a 24-hour period, it has only been possible to collect link travel times for the $t + 1$ time-step (i.e.,\ next 15 minutes). Travel times for each link were collected at a 15-minute interval between 06:00 and 22:00.

Another important aspect is that the Google Traffic model is primarily designed for estimating car travel times, and therefore it can be biased and not ideal for estimating the bus travel times used in our experiments. Since we only consider link travel times and collect data for each link individually, the bus dwell time will not be an issue, as it is not included in either measurement.

\begin{figure*}
  \centering
  \includegraphics[width=\textwidth]{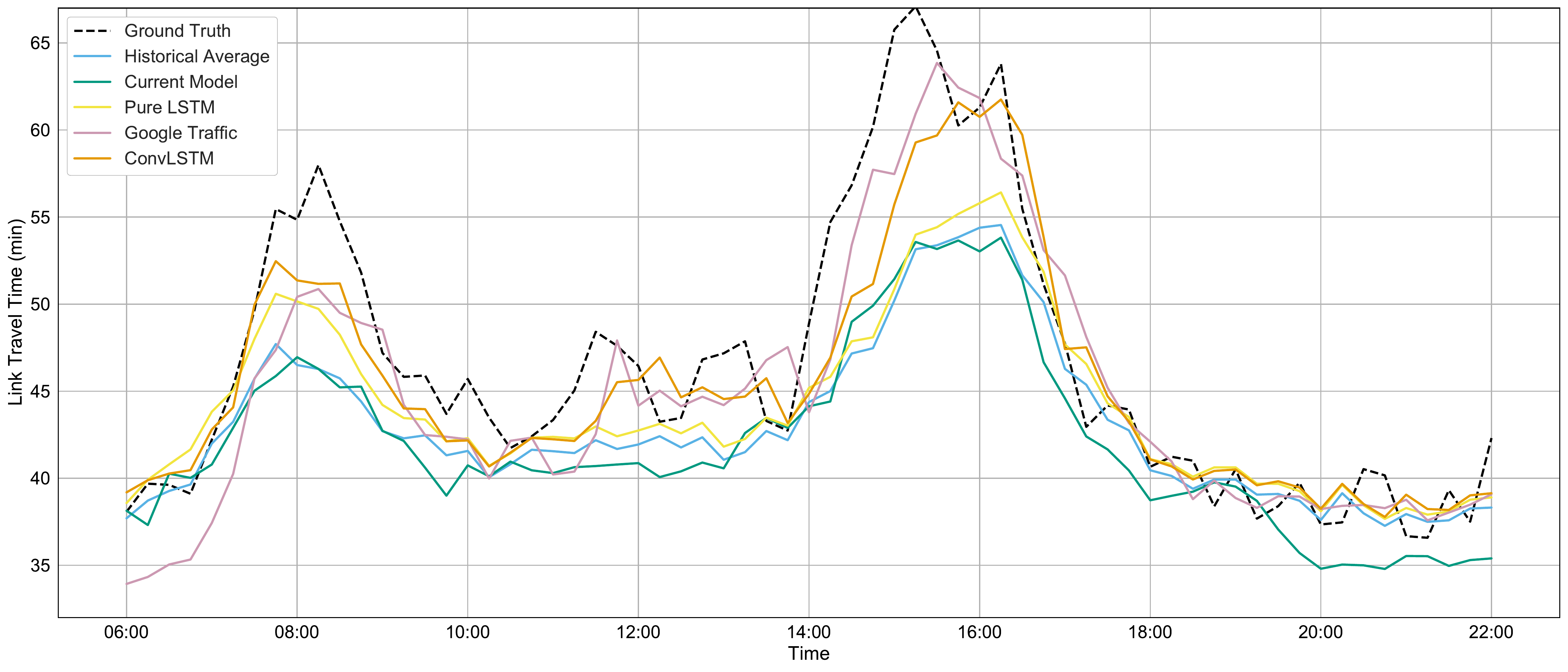}
  \caption{Accumulated link travel time over a single day (a Thursday) in the test set.}
  \label{fig:comparison_day}
\end{figure*}
\begin{figure*}
  \centering
  \includegraphics[width=\textwidth]{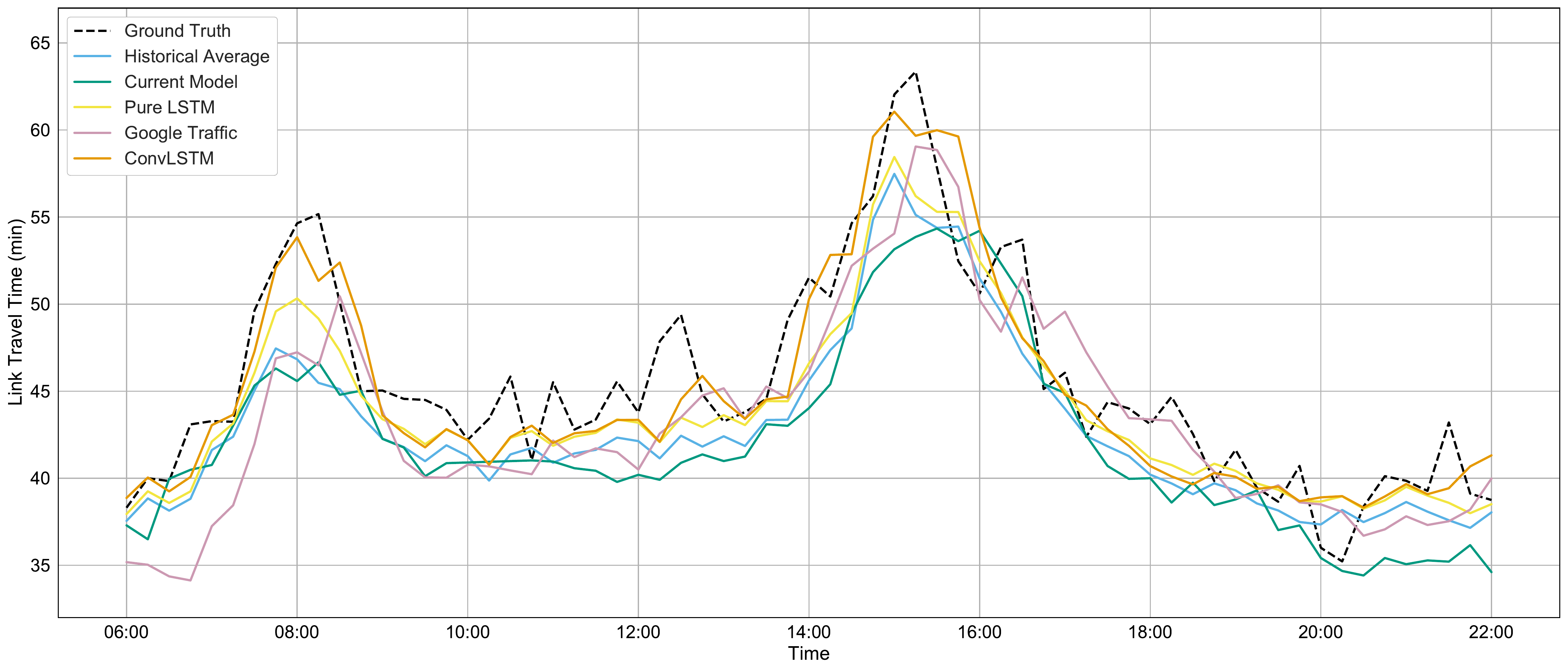}
  \caption{Accumulated link travel time over a single day (a Friday) in the test set.}
  \label{fig:comparison_day_2}
\end{figure*}

\subsection{Comparison}
We compare the performance of the proposed \emph{ConvLSTM} model for bus link travel time prediction against the baseline models mentioned above. The overall results from \Cref{tab:results} show that the \emph{ConvLSTM} model outperforms all the other methods. The \emph{current model} performs the worst, even compared to the \emph{historical average} model, on which it is based on. This is most likely due to the rule-based enforcement put on top of the historical average.

Although the difference in performance might seem small, it should be emphasized that the evaluation measurements are averaging their errors, and thus the increased accuracy can be much higher on individual journeys, especially if they experience very irregular travel times. To investigate this, we focus our analysis on periods when the transport system is most vulnerable, and even small changes in regularity can propagate, since recovery is not an option, i.e.\ during morning and afternoon peaks.

\begin{table}[!ht]
  \center
  \begin{tabular}{l|rrr}
  Model &  RMSE &  MAE &  MAPE \\
  \hline
  \hline
  Historical Average &        6.40 &       5.57 &     10.62 \% \\ \hline
  Current Model      &        6.69 &       5.88 &     11.22 \% \\ \hline
  Pure LSTM          &        3.80 &       3.16 &      6.01 \% \\ \hline
  Google Traffic     &        5.25 &       4.62 &      9.17 \% \\ \hline
  ConvLSTM           &        2.64 &       2.09 &      4.04 \% \\ \hline
  \end{tabular}
  \caption{Results: Morning peak (7h--9h)}
  \label{tab:morning_peak}
  
\end{table}

\begin{table}[!ht]
  \center
  \begin{tabular}{l|rrr}
  Model &  RMSE &  MAE &  MAPE \\
  \hline
  \hline
  Historical Average &    5.90 &   4.65 &  8.28 \% \\ \hline
  Current Model      &    6.28 &   5.20 &  9.37 \% \\ \hline
  Pure LSTM          &    5.26 &   3.97 &  7.08 \% \\ \hline
  Google Traffic     &    4.16 &   3.34 &  6.21 \% \\ \hline
  ConvLSTM           &    3.79 &   3.02 &  5.61 \% \\ \hline
  \end{tabular}
  \caption{Results: Afternoon peak (14h--18h)}
  \label{tab:afternoon_peak}  
\end{table}

Tables~\ref{tab:morning_peak} and \ref{tab:afternoon_peak} show the evaluation results for \emph{morning peaks (weekdays, 7h--9h)} and \emph{afternoon peaks (weekdays, 14h--18h)}, respectively, for the time-step $t + 1$.

The peak hour evaluation shows that the \emph{ConvLSTM} model increases its performance over the baseline models when the transport network is put under stress. In the morning peak, the \emph{ConvLSTM} model does not degrade in performance compared to the overall daytime results, whereas all the baseline models experience a decrease in performance of up to several minutes according to both RMSE and MAE, and an increase in MAPE of roughly one third. 

Similarly, the afternoon peak evaluation shows improvements with respect to the baseline models, even though the \emph{ConvLSTM} model also decreases its performance when compared to the overall results. However, in this case, the difference in performance with the baseline methods is not as significant as in the morning peak. We can also observe that the \emph{Google Traffic} model performs rather well in the afternoon peak, which reduces the gap in error to less than a minute to the proposed ConvLSTM-based approach. 

To obtain a more detailed view of how the different models perform at the micro-level (i.e.\ the specific journey), we can inspect a single day of predictions. A random weekday from the test dataset is plotted in~\Cref{fig:comparison_day} which shows the accumulated travel time of all 32 links and the predicted travel time at time-step $t + 1$, both for the proposed model and the baseline model.

On this particular day (a Thursday), the peak hour traffic was worse than normal, which leads both the \emph{historical average} model and the \emph{current model} to underestimate travel time in the peak periods. Please recall that the \emph{current model} is based on the \emph{historical average} model. Therefore, it is not surprising that they perform similarly. There is also a small peak in travel time in the afternoon, which none of the historical average models is able to predict.

On the other hand, both the \emph{Google Traffic} model and the proposed \emph{ConvLSTM} model get much closer to the ground truth in the peak hours. The \emph{Google Traffic} model seems to predict more accurately than the \emph{ConvLSTM} model in the afternoon peak, whereas the opposite occurs in the morning peak. However, both models are able to detect the irregular peak in the afternoon and adjust to it, at least to some degree.

\Cref{fig:comparison_day_2}~shows another example day - a Friday. Here the difference between the proposed model and the historical average and current model baselines is slightly less significant, simply because the day to a larger degree follows the average pattern for a ``normal'' Friday (especially around the afternoon peak). Nonetheless, the proposed model still performs the best, and this also supports our claim that the proposed model is strongest when the traffic pattern deviates from the normal pattern, i.e.\ when the transport network is under stress.

Finally, we compare the computational complexity of training the different models. Obviously, we cannot include metrics for the \textit{Google Traffic}, as the model is not public. Likewise, it is not sensible to compare with the \textit{Current Model}, since it is ``trained'' on a dataset of different size and on hardware using in production at the transport authority. But, since we know it is essentially an \textit{historical average} approach, we can expect a similar computational complexity. The historical average can be calculated within seconds for the full 23-week training dataset. The training of the \textit{Pure LSTM} and \textit{ConvLSTM} model can be achieved in both cases, for the full 23-week training dataset and the full 32-links, in less than 20 minutes on commodity hardware (8 cores, 64 GB RAM, GTX 1070 GPU). This might indicate why the \textit{historical average} models are still popular in the industrial systems, but we, however, argue that the more complex models are indeed scalable and the improved accuracy desirable, even though it is more computationally expensive.

\section{Conclusion}
\label{sec:conclusion}
This paper proposed a multi-output, multi-time-step system for bus travel time prediction. The proposed system uses a deep neural network model consisting of convolutional and long short-term memory (LSTM) layers, that is able to capture the non-static spatio-temporal correlations of variability in urban bus travel times. This allows the model to generalize patterns learned in predictions across space and time. Also, our approach for multi-time-step prediction using an encoder/decoder architecture is, to the best of our knowledge, new in the context of bus travel time prediction. The proposed approach allows accurate predictions further into the future compared to traditional approaches where subsequent time-steps are predicted independently. Our empirical results demonstrate that the proposed model outperforms other popular and recent methods from the state-of-the-art. This includes Google's Traffic model based on crowd-sourced live traffic data, and the current model deployed by Movia, the public transport authority in the Greater Copenhagen Area. The increased accuracy when compared to the baseline approaches is even more significant in the peak hours, where the urban bus transport network is under stress. The data required for the proposed system is simply the standard output that most AVL systems used in the public transport industry produce. We are aware that public transport agencies in Singapore, London, New York, Stockholm, Oslo, and Helsinki all have deployed AVL systems that fulfill this requirement, and thus the proposed system indeed generalizes trivially across different cities in different countries. 

Although the proposed method is more computationally expensive than simple \textit{historical average} models, given the state of modern computational hardware, it is indeed scalable to be applied to an urban bus network for independent routes. Even with commodity hardware, we are able to retrain the route used in this experiment in less than 20 minutes, and we can thus easily retrain the model on a daily basis. Given the results of our proposed model, we are currently actively pursuing deployment of the model in the Greater Copenhagen region, in close collaboration with the transport authority - Movia. We do however consider this route-independent approach a limitation of the current system, and below we provide some research opportunities to extend the proposed system by handling correlations between different routes. 

\subsection{Future work}
As future work, we would like to extend the presented systems in the following directions:
\begin{enumerate}[i)]
    \item The integration of our proposed system to different control strategies for enhancing the regularity and reliability of the bus service, e.g.\ as suggested by \citet{Lo2012}. This would create a possible feedback loop from the predicted travel times that could possibly affct the travel times on a short-term basis. This is a non-trivial task, since it requires either simulation, which is complex for urban public transport networks in the detail needed here, or integration directly into currently running services, which is organizationally and technically challenging.
    \item In order for the prediction accuracy to be increased further, it would be interesting to include more contextual features in the input data and not only the observed travel times. This could include features from the road network that the link consists of, e.g. whether intersections on the link are regulated by a traffic signal or not. In order to achieve this, map matching of at least the link geometry to the road network is necessary. However, this should be easily overcome, and many interesting crowd-sourced data are freely available (e.g.\ Open Street Map). Additional data sources such as weather conditions have shown to impact bus travel time \citep{Chen2004} and could also be included. More rare, but highly impacting deviations, such as traffic incidents, service-outage, holidays and large events, could also be considered in this research direction.
    \item Currently, the model only uses convolutions over a single bus route, i.e.\ 1D-convolutions. We believe that it would be interesting to see the effect on the accuracy of the system if this was generalized to a network of bus routes. This would require extending the convolutions into a multi-dimensional space. Popular approaches used traditionally in conjunction with convolutions, such as overlaying the geographical map with a 2D-grid, have not shown good results. The challenge seems to be that bus networks are relatively sparse, and that travel times do not aggregate well in cells, e.g. compared to travel demand. Recent state-of-the-art proposes \textit{graph convolutional neural networks} \citep{Li2017}, i.e.\ where the convolutions are done over graph structures. We plan to pursue this approach - with the complications and development needed - to adapt the method to bus networks and the bus travel time prediction problem.
    \item A final direction we have identified is to include the proposed system in an ensemble/multi-model approach. In this case, the proposed model can be included and used as a sub-model for the ensemble. The challenge here is the coordination between the different sub-models that can be seen as autonomous agents in an expert and intelligent system context. Especially, how to solve disagreements. Different approaches have been proposed by \citet{Weng2018}, and we expect to explore these approaches in our research.
\end{enumerate}

\bibliography{library}

\end{document}